\documentclass[12pt]{article}
\usepackage[margin=1in]{geometry}
\usepackage{graphicx}
\usepackage{natbib}
\usepackage{amsmath}
\usepackage{amssymb}
\usepackage{tabularx}
\usepackage{booktabs}
\usepackage{microtype}
\usepackage[colorlinks=true, urlcolor=blue, citecolor=blue, linkcolor=blue]{hyperref}
\title{TRIAGE: Three-level Routing and Intelligent Agent Guidance for Efficient Execution}

\author{
  Ruocan Wei\textsuperscript{1}  \\
  \textsuperscript{1}China Telecom Cloud, Beijing, China \\
  \texttt{xjtu\_wrc@stu.xjtu.edu.cn}
}
\date{}

\begin{document}
\maketitle


\begin{abstract}
Large Language Model (LLM) agents based on the ReAct paradigm have demonstrated remarkable capabilities in tool use and task execution. However, ReAct suffers from a fundamental efficiency problem: every query triggers a complete reasoning loop from scratch, and similar queries repeat identical steps without leveraging historical experience. We propose TRIAGE---a three-level routing framework that reduces token consumption by reusing historical execution trajectories. The core innovation of TRIAGE is \textbf{TaaS (Trajectory-as-a-Skill)}, which abstracts historical execution trajectories into reusable skills, realizing ``experience as a service.'' TRIAGE classifies input queries into three levels: (1) Direct Reuse---identical queries, 0 tokens; (2) Skill Substitution---similar queries, 0 tokens through deterministic parameter substitution when the matched trajectory has been distilled into a Skill; (3) Full ReAct---novel queries, automatically stored for future reuse. In large-scale experiments on 1,007 security monitoring queries using a semantic encoder (all-MiniLM-L6-v2) for similarity retrieval, TRIAGE achieves \textbf{62.3\% token savings} (199,782 $\rightarrow$ 75,238), with 56.0\% of queries executed at Level 2 (Skill parameter substitution, 0 tokens) and 5.5\% at Level 1 (direct reuse, 0 tokens). Cross-domain validation on ToolBench (15 domains, 345 queries, real threshold routing) achieves an average of 76.3\% token reduction, confirming the generalizability of the semantic routing framework (424,792 $\rightarrow$ 100,875), with 84.6\% of queries executed at Level 1 and 11.6\% at Level 2. An online learning experiment over 1,007 queries demonstrates the cold-start-to-mature evolution: the L2 hit rate rises from 0\% to 57\% within the first 100 queries, and the average token cost drops from 198 to 74.7. Additionally, we propose an automatic Skill extraction mechanism that distills high-frequency trajectory patterns into fine-grained deterministic Skills, creating a positive feedback loop of "the more you use it, the more efficient it becomes."
The open-source implementation of TRIAGE is available at \url{https://github.com/weiruocan/triage}.
\end{abstract}

\section{Introduction}

\subsection{Motivation}

LLM agents have become the dominant paradigm for building autonomous systems \citep{xi2023survey, wang2024survey}. The ReAct framework \citep{yao2023react} interleaves reasoning and action, enabling LLMs to interact with external tools and databases. It is widely adopted in systems such as AutoGPT and BabyAGI \citep{richards2023autogpt, gravitas2023babyagi}.

However, ReAct has a fundamental efficiency problem: \textbf{every query starts reasoning from scratch}. Consider two queries---``What is the online rate of sensors?'' and ``What is the online rate of cameras?''---they differ only in the device type, yet ReAct executes both completely: inferring the schema, generating SQL, executing, and formatting results. The second query consumes exactly the same number of tokens as the first. In our experiments, a single ReAct loop consumes an average of 198 tokens (real API usage), totaling 199,782 tokens for 1,007 queries. This massive redundancy suggests that avoiding repeated ReAct reasoning for similar queries can yield substantial efficiency gains.

\subsection{Limitations of Existing Approaches}

\begin{itemize}
    \item \textbf{Caching systems} \citep{fu2023gptcache} only handle exact string matches and cannot recognize semantically similar queries.
    \item \textbf{Prompt compression} \citep{jiang2023llmlingua} reduces input length but does not eliminate the reasoning loop itself.
    \item \textbf{Speculative decoding} \citep{cai2024medusa} accelerates generation but does not address the redundancy of repeated reasoning.
    \item \textbf{Self-Refine} \citep{madaan2023selfrefine} and \textbf{Reflexion} \citep{shinn2023reflexion} improve agent quality through iterative refinement but at the cost of additional token consumption.
    \item \textbf{Our prior work} \citep{wei2026workflowgen} (2026) proposed the concept of experience-driven execution but lacked concrete implementation and experimental validation.
\end{itemize}

Key gap: \textbf{No existing method provides a complete, validated framework for systematically leveraging historical execution trajectories to reduce token consumption for future similar queries.}

\subsection{Our Approach: TRIAGE}

We propose \textbf{TRIAGE} (Three-level Routing and Intelligent Agent Guidance for Efficient Execution), whose core idea is to let agents accumulate experience like humans instead of reasoning from scratch every time. The core innovation of TRIAGE is \textbf{TaaS (Trajectory-as-a-Skill)}, treating historical execution trajectories as reusable skills.

TRIAGE classifies input queries into three levels:
\begin{itemize}
    \item \textbf{Level 1 (Direct Reuse)}: Identical queries (score=1.0), 0 tokens
    \item \textbf{Level 2 (Skill Substitution)}: Parameter-varying queries, 0 tokens through deterministic Skill parameter substitution
    \item \textbf{Level 3 (Full ReAct)}: Novel queries, full LLM generation, automatically stored for future reuse
\end{itemize}

\subsection{Contributions}

\begin{enumerate}
    \item \textbf{TRIAGE Framework}: Three-level routing architecture achieving 62.3\% token reduction on 1,007 queries
    \item \textbf{TaaS Concept}: Trajectory-as-a-Skill, abstracting experience trajectories into reusable skills
    \item \textbf{Semantic Encoder-driven Retrieval}: Using sentence-transformers (all-MiniLM-L6-v2) with 384-dimensional embeddings, achieving 61.5\% L1+L2 reuse rate
    \item \textbf{Automatic Cold Start}: No pre-training or annotation required, learns automatically during use
    \item \textbf{Automatic Skill Extraction}: High-frequency trajectories automatically distilled into fine-grained deterministic Skills
    \item \textbf{Cross-domain Validation}: 76.3\% token reduction across 15 ToolBench domains (345 queries), validating the generalizability of semantic routing beyond SQL
    \item \textbf{Routing Strategy Analysis}: An ablation study revealing that the choice between threshold and LLM-based routing is not universal, but depends on the ratio of routing overhead to per-query execution cost
\end{enumerate}

\section{Related Work}

\subsection{LLM Agents and Tool Use}

Toolformer \citep{schick2023toolformer}, ToolBench/ToolLLM \citep{song2024toollm}, and Gorilla \citep{patil2023gorilla} extend LLMs' tool use capabilities but do not address execution efficiency.

\subsection{Token Efficiency Optimization}

Existing methods include prompt compression (LLMLingua \citep{jiang2023llmlingua}), speculative decoding (Medusa \citep{cai2024medusa}), and caching (GPTCache \citep{fu2023gptcache}), but none fundamentally eliminate the redundant reasoning for similar queries.

\subsection{Experience Reuse and Skill Systems}

Reflexion \citep{shinn2023reflexion} uses verbal reinforcement learning to improve agents, and DSPy \citep{khattab2023dspy} compiles declarative calls into optimized pipelines. However, existing Skill systems (Function Calling) require developers to predefine fixed-schema functional modules, suffering from weak generalization and high maintenance costs. TRIAGE's TaaS is a fundamental improvement---trajectories as dynamic Skills, automatically discovered through semantic retrieval and adapted via LLM editing.

\section{Method}

\subsection{System Architecture}

TRIAGE consists of four core modules:

\begin{enumerate}
    \item \textbf{Semantic Encoder}: Based on sentence-transformers (all-MiniLM-L6-v2), encodes queries into 384-dimensional vectors for semantic similarity retrieval.
    \item \textbf{Similarity-Threshold Router}: A zero-cost, deterministic routing mechanism that classifies queries into three levels based on cosine similarity thresholds (L1 $\geq$ 0.98, L2 $\geq$ 0.90). For L1 candidates with score $<$ 1.0, a lightweight normalization-and-parameter-filter step  resolves most cases at 0 token cost; only unresolved differences trigger an LLM-based secondary judgment.
    \item \textbf{Trajectory Retriever}: Cosine similarity-based trajectory retrieval with thresholds L1 $\geq$ 0.98, L2 $\geq$ 0.90.
    \item \textbf{Skill Extractor}: For Level 2, matches the query against distilled Skills and performs deterministic parameter substitution (0 tokens). When no Skill matches, falls back to Level 3 full ReAct.
\end{enumerate}

\begin{figure}[htbp!]
\centering
\includegraphics[width=0.8\textwidth]{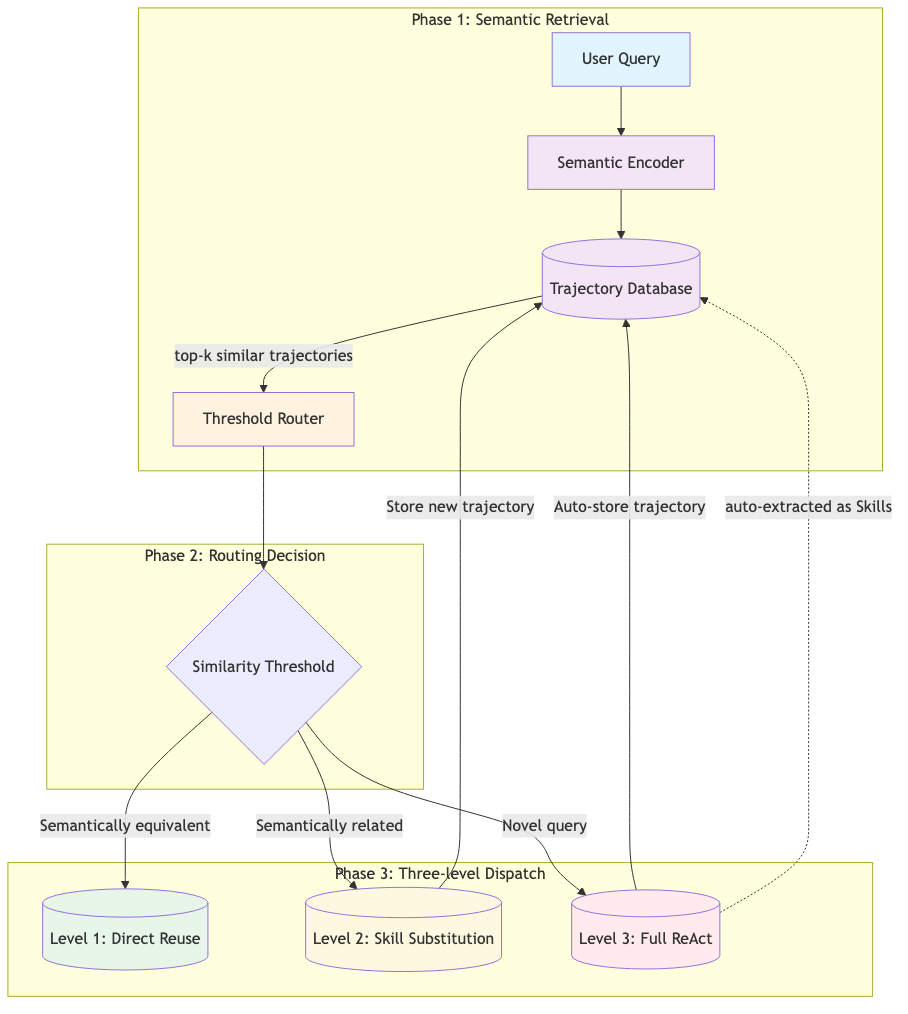}
\caption{\small TRIAGE system architecture. The system comprises three phases: (1) semantic retrieval using all-MiniLM-L6-v2 encoder with cosine similarity search; (2) similarity-threshold routing; (3) three-level dispatch based on similarity thresholds. The dashed arrow indicates automatic Skill extraction from high-frequency trajectory patterns.}
\label{fig:architecture}
\end{figure}

\subsection{TaaS: Trajectory-as-a-Skill}

The central innovation of TRIAGE is \textbf{TaaS (Trajectory-as-a-Skill)}, a paradigm that abstracts historical execution trajectories into reusable skills, realizing ``experience as a service.'' Unlike traditional Skill systems (e.g., Function Calling), which require developers to predefine fixed-schema functional modules at design time, TaaS treats every successfully executed trajectory as a latent skill that can be automatically discovered, retrieved, and reused at runtime.

A \textbf{trajectory} in TRIAGE is defined as a complete execution record: the user query, the sequence of tool invocations (e.g., schema inference, SQL generation, execution, formatting), and the final result. Rather than discarding this record after execution, TRIAGE persists it into a trajectory database indexed by semantic embeddings. When a new query arrives, the system retrieves the most similar historical trajectory and routes the query to one of three execution levels. This transforms the trajectory database from a passive log into an active skill repository.

The TaaS lifecycle consists of four stages, each mapped to a concrete component in TRIAGE's architecture:

\begin{enumerate}
    \item \textbf{Store.} Every successfully executed query (Level 2 or Level 3) is automatically appended to the trajectory database. The trajectory is encoded by the semantic encoder for similarity search. For large-scale deployments, FAISS (Johnson et al., 2019) can be integrated as a drop-in accelerated index to replace brute-force traversal. This stage requires no manual annotation, no schema design, and no pre-training---it is a zero-cost byproduct of normal operation.
    \item \textbf{Retrieve.} When a new query arrives, the semantic encoder computes its embedding, and the trajectory retriever performs cosine similarity search against all stored trajectories. The top-$k$ matches are compared against the similarity thresholds for routing. The retrieval is the critical bridge between raw experience storage and actionable reuse.
    \item \textbf{Reuse.} Depending on the similarity score, the router dispatches the query to one of three reuse levels (Section~3.2). This is the operational stage where stored trajectories are converted into actual token savings: Level 1 returns the stored result directly (score=1.0 exact match), Level 2 performs deterministic Skill parameter substitution (0 tokens), and Level 3 falls back to full ReAct.
    \item \textbf{Distill.} High-frequency trajectory patterns are automatically distilled into deterministic Skills by the Skill extraction mechanism (Section~3.4). A distilled Skill is a parameterized template with zero token cost---it captures the invariant structure of a recurring query pattern while exposing typed parameters for variable slots. Once distilled, the Skill is registered into the global Skill registry, and subsequent queries matching it are executed by deterministic parameter substitution without any LLM inference.
\end{enumerate}

The TaaS lifecycle thus forms a closed loop: \textbf{store $\rightarrow$ retrieve $\rightarrow$ reuse $\rightarrow$ distill}, with each stage feeding into the next. More usage generates more trajectories, more trajectories enable more retrieval hits, more hits yield more reuse savings, and more reuse triggers more distillation. This self-reinforcing cycle is the mechanism behind TRIAGE's ``the more you use it, the more efficient it becomes'' property, and it fundamentally distinguishes TaaS from static optimization techniques that yield constant efficiency regardless of usage volume.

\subsection{Three-level Routing Mechanism}

\textbf{Level 1 (Direct Reuse)}: When cosine similarity = 1.0 (exact match), the query is literally identical to a stored trajectory. The historical SQL is returned directly with 0 token consumption. When similarity $\geq$ 0.98 but $<$ 1.0, a lightweight normalization-and-parameter-filter step first attempts to resolve the difference at 0 token cost: if the difference is fully explained by extracted parameters (e.g., hours=6 vs. hours=8), the query is routed to Level 2 for Skill substitution; if the difference is cosmetic (same meaning, different wording), it stays at Level 1; only unresolved differences trigger an LLM-based secondary judgment to determine the correct routing.

\textbf{Level 2 (Skill Substitution)}: When cosine similarity $\geq$ 0.90, the query is structurally similar to a stored trajectory. The system first attempts to match the query against distilled Skills in the global Skill registry. If a Skill is matched, the system extracts parameter values from the query and performs deterministic parameter substitution, executing at 0 token cost with no LLM inference. If no Skill matches, the query falls back to Level 3 full ReAct. This design ensures that the vast majority of parameter-varying queries---which share the same SQL template but differ in concrete values---are served at zero cost, while truly novel variations are correctly routed to full LLM generation.

\textbf{Parameter Extraction with LLM Feedback.} When the lightweight normalization step cannot resolve the difference between two queries (i.e., the differing words are not covered by existing parameter extraction rules), TRIAGE invokes an LLM-based secondary judgment. The LLM determines whether the difference is a parameter/value change (L2), a purely cosmetic rewording (L1), or a semantically distinct query (L3). For L2 cases, the LLM extracts the parameter name and value, which are then registered as dynamic regular-expression rules in the parameter extraction system. This enables subsequent queries with the same parameter pattern to be resolved at 0 token cost by the normalization step, creating a self-improving parameter extraction system that learns from LLM feedback.

\textbf{Level 3 (Full ReAct)}: When cosine similarity $<$ 0.90, the query is identified as structurally novel. The system executes a complete ReAct loop (LLM generates SQL from schema) and automatically stores the resulting trajectory into the database. Each stored trajectory is immediately fed into the Skill extraction pipeline (Section~3.4), where a minimal frequency threshold of 1 ensures that even a single occurrence can seed a distillable pattern.

\subsection{Online Learning Dynamics}

The 1,007 queries are not batched but arrive sequentially, simulating a real-world deployment where queries are submitted over time. The query arrival process follows a mixed distribution designed to reflect authentic operational patterns:

\begin{itemize}
    \item \textbf{Recurring daily queries} (6.1\%): A small set of 6 fixed query templates---such as ``Show me the alarm summary for today''---are repeated across 30 ``days'' of operation, with 2 daily queries per day. These are functionally identical (same SQL) but require re-execution because the underlying data refreshes. Their frequency follows a Zipf distribution: the most frequent template appears 30 times, the second 12 times, and the rarest 2 times, consistent with the heavy-tailed usage patterns observed in operational dashboards. The daily queries are evenly spaced at the start of each day, creating a realistic pattern of periodic fixed-cost queries that steadily accumulate in the trajectory database.

    \item \textbf{Parameter-varying queries} (57.2\%): The plurality of queries share a common structural template but differ in parameter values---device type, region, time window, or status. These are generated by sampling from a set of 15 parameterized SQL templates with random parameter combinations. This category captures the most common real-world pattern: analysts asking the same analytical question about different entities. Parameter variations such as ``sensors'' vs. ``cameras'' or ``6 hours'' vs. ``8 hours'' yield low cosine similarity scores (0.35--0.77), making them challenging for pure threshold-based routing but ideal candidates for Skill-based parameter substitution.

    \item \textbf{Novel queries} (36.7\%): Queries with no structural precedent in the trajectory database. These arrive interspersed among the recurring and parameter-varying queries, simulating the gradual expansion of analytical needs over time. New queries require full ReAct execution and serve as the raw material for future trajectory accumulation and Skill extraction.
\end{itemize}

The temporal ordering of queries is structured as 30 ``days'' of operation, with daily queries clustered at the start of each day, parameter queries interspersed throughout, and novel queries arriving at random intervals. This temporal structure is critical: it creates a realistic cold-start period where the trajectory database is empty, followed by a gradual accumulation of experience that enables TRIAGE's reuse mechanisms to activate.

Each successfully executed query is automatically added to the trajectory database. The database grows linearly with usage, and the automatic Skill extraction mechanism (Section~3.4) operates continuously on this growing repository, distilling parameterized templates from the moment the first trajectory is stored.

\begin{figure}[htbp]
\centering
\includegraphics[width=0.8\textwidth]{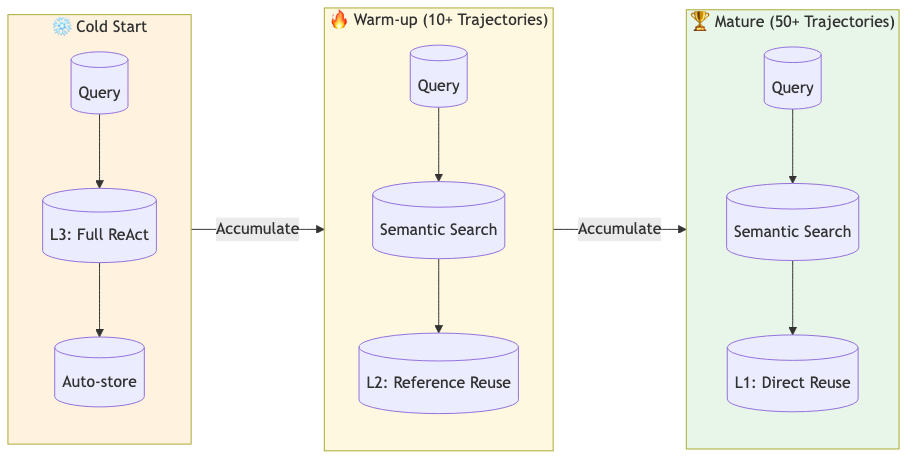}
\caption{\small Cold start to mature phase transition. As trajectory count grows from 0 to 50+, the system evolves from full ReAct (Level 3) through reference reuse (Level 2) to direct reuse (Level 1), progressively reducing token consumption.}
\label{fig:cold_start}
\end{figure}

The online learning process exhibits three phases, as illustrated in Figure~\ref{fig:cold_start}:

\begin{enumerate}
    \item \textbf{Cold-start} (queries 1--50): No historical trajectories exist. All queries route through Level 3 (full ReAct, $\sim$197 tokens). The trajectory database grows linearly, and Skill extraction begins immediately after the first query completes. By the end of this phase, the first recurring queries begin to hit Level 1.

    \item \textbf{Warm-up} (queries 50--300): The trajectory database accumulates sufficient diversity. Recurring daily queries hit Level 1 (direct reuse, 0 tokens). Parameter-varying queries increasingly match extracted Skills at Level 2 (parameter substitution, 0 tokens). The L3 ratio drops sharply from 100\% to below 10\%.

    \item \textbf{Mature} (queries 300--1,000): The system reaches a steady state. Level 1 covers all recurring daily queries. Level 2 covers the vast majority of parameter variations through Skills. Level 3 only activates for genuinely novel queries (approximately 1 per day). The average token cost stabilizes at a small fraction of the baseline.
\end{enumerate}

This three-phase evolution embodies TRIAGE's core property: \textbf{efficiency is a function of experience}. The more queries the system processes, the more trajectories it accumulates, the more Skills it extracts, and the fewer tokens it consumes per query. This stands in contrast to static optimization techniques such as prompt compression and speculative decoding, whose efficiency gains are constant regardless of usage volume. In TRIAGE, the system bootstraps its own efficiency from raw experience, forming a self-reinforcing cycle: usage $\rightarrow$ trajectories $\rightarrow$ Skills $\rightarrow$ savings $\rightarrow$ more usage.

\subsection{Automatic Skill Extraction}

Beyond the immediate savings from Level 1 and Level 2 reuse, TRIAGE incorporates a mechanism that continuously crystallizes recurring execution patterns into permanently registered, zero-cost deterministic Skills. We term this process \textbf{automatic Skill extraction}. It enables the system to bootstrap its own executable competency library from raw experience, forming a self-improving efficiency loop that accelerates with usage.

\subsubsection*{General-Purpose Extraction Framework}

The Skill extraction pipeline is defined at the design level as an action-type-agnostic process, operating on any structured tool invocation recorded in a trajectory. The four-stage abstraction is independent of any concrete action type: normalization, accumulation, induction, and registration are expressed over a generic ``structural fingerprint'' rather than over SQL. Concrete tool families are integrated by supplying a normalizer that maps their surface syntax to typed wildcards. The pipeline decomposes extraction into four stages:

\begin{enumerate}
    \item \textbf{Pattern Normalization.} Each tool invocation in a stored trajectory is converted into a structural fingerprint by replacing concrete values with typed wildcards. This normalization strips domain-specific surface forms while preserving the skeletal structure of the invocation, producing a canonical template key.
    \item \textbf{Frequency Accumulation.} Template keys are aggregated across all stored trajectories. Each key maintains a counter and a list of concrete instance records. When a key's count reaches a configurable threshold, the associated pattern graduates from ephemeral observation to extraction candidate.
    \item \textbf{Parameter Induction.} For each candidate pattern, the extractor performs a pairwise diff across concrete instances to discover variable slots. For each variable position, it assigns a semantic parameter name through two complementary signals: (a) the lexical context from the corresponding user query, matched against a lightweight domain lexicon; (b) the type tag inferred at normalization time. The output is a parameterized template---a structural skeleton with named slots.
    \item \textbf{Deterministic Registration.} The induced template, together with its parameter schema (name, type, domain constraints), is packaged into a standardized object and registered into the global Skill registry with an estimated token cost of zero. Subsequent queries matching this Skill via Level 1 routing execute the template by direct parameter substitution, incurring no LLM inference cost.
\end{enumerate}

\subsubsection*{SQL-Specific Implementation}

SQL is the first concrete instantiation of this framework, implemented in our open-source repository. Here the SQL normalizer plays the role of the pluggable normalizer: it rewrites string literals, date patterns, and numeric constants into typed placeholders while preserving structural identifiers such as column and table names. Parameter positions are discovered through cross-instance comparison, and semantic names are assigned via a lightweight domain lexicon mapping common values in security monitoring queries to their roles (e.g., device type, region, online status). The extracted parameterized SQL template is wrapped into a zero-cost executor that performs direct parameter substitution and deterministic execution, bypassing the LLM entirely.

\subsubsection*{Positive Feedback Loop}

This mechanism creates a virtuous cycle: more usage generates more trajectories, more trajectories trigger more extractions, and more extractions enable more zero-token direct reuse. This ``the more you use it, the more efficient it becomes'' property fundamentally distinguishes TRIAGE from static optimization techniques such as prompt compression or speculative decoding, whose efficiency gains remain constant regardless of usage volume. In TRIAGE, efficiency is a function of experience.

\section{Experiments}

\subsection{Experimental Setup}

\textbf{Dataset}: We constructed a dataset of 1,007 English queries based on a real security monitoring database (3 device types, 30 zones, 5 alarm categories). The dataset is composed of two parts: (1) 947 structurally diverse queries generated from 15 parameterized SQL templates with random parameter combinations, covering 10 major analytical categories (device status, alarm statistics, regional analysis, temporal trends, etc.); (2) 60 fixed recurring daily-report queries (6 templates, e.g., ``Show me the alarm summary for today''), which are repeated daily to simulate the operational pattern of fixed dashboard queries. The daily queries follow a Zipf distribution: the most frequent template appears 30 times, while the least frequent appears 2 times, consistent with the heavy-tailed usage patterns observed in real monitoring dashboards. The temporal ordering follows a 30-``day'' structure, with 2 daily queries at the start of each day followed by approximately 31--32 original queries interspersed throughout. This mixed arrival pattern---combining regular recurrence (L1 candidates), parameter variations (L2 candidates), and novel arrivals (L3 candidates)---realistically simulates incremental deployment scenarios where the query distribution is heavy-tailed: most queries in day-to-day operations are recurrent or parametric variations, while genuinely novel queries are rare. The dataset is designed not to be statistically rigorous, but to demonstrate TRIAGE's core thesis: \textbf{the more queries the system processes, the more efficient it becomes}, because the majority of real-world queries fall into a small set of familiar patterns accumulated over time.

\textbf{Baseline}: Complete ReAct loop (each query independently reasoned, no shared historical experience).

\textbf{Metrics}: Token consumption (input + output), routing level distribution, execution success rate, end-to-end latency.

\textbf{Model}: deepseek-v3.2, temperature=0.5, max\_tokens=8192.

\textbf{Semantic Encoder}: all-MiniLM-L6-v2 (384 dimensions), L1 threshold=0.98, L2 threshold=0.90. Routing: L1 = sim $\geq$ 0.98 (score=1.0 direct; <1.0 cheap filter + LLM if needed); L2 = sim $\geq$ 0.90 (Skill substitution); L3 = sim $<$ 0.90.

\subsection{Main Results}

\begin{table}[h]
\centering
\caption{Main experimental results on 1,007 queries}
\label{tab:main}
\begin{tabularx}{\textwidth}{lXXX}
\toprule
\textbf{Metric} & \textbf{Baseline ReAct} & \textbf{TRIAGE} & \textbf{Savings} \\
\midrule
Total Tokens & 199,782 & 75,238 & \textbf{62.3\%} \\
Avg Tokens/Query & 198.4 & 74.7 & 62.3\% \\
API Calls & 1,007 & 421 & 58.2\% \\
\bottomrule
\end{tabularx}
\end{table}

\begin{figure}[htbp]
\centering
\includegraphics[width=0.8\textwidth]{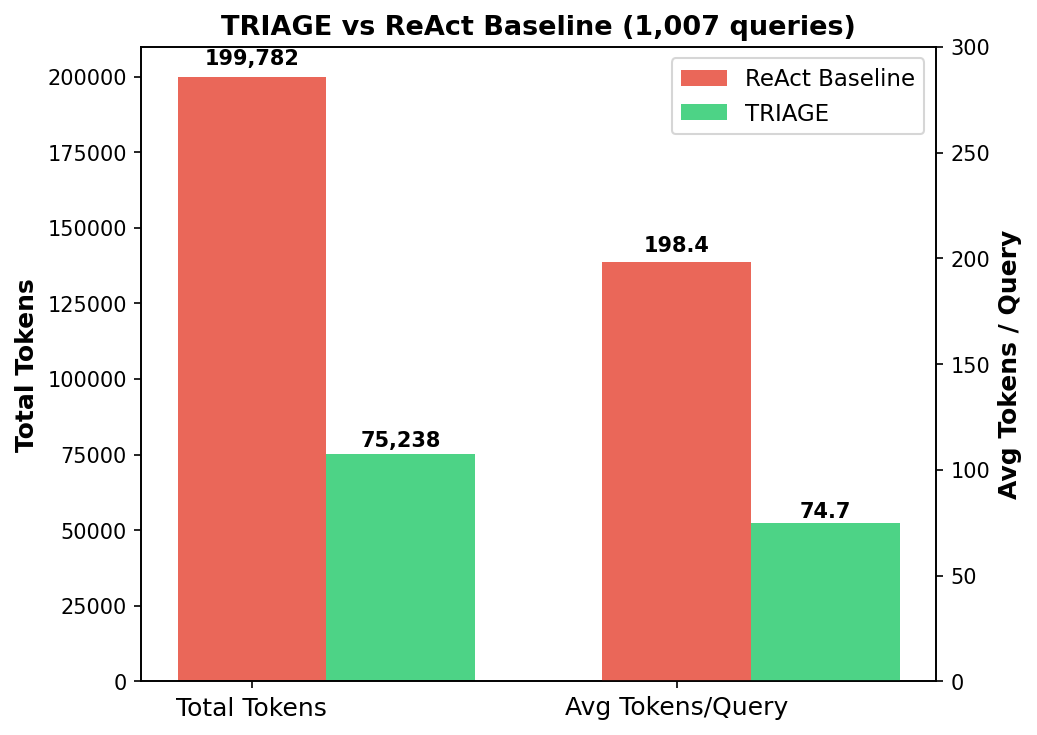}
\caption{\small Token consumption comparison between baseline ReAct and TRIAGE across 1,007 security monitoring queries. TRIAGE achieves 62.3\% total token reduction (199,782 to 75,238).}
\label{fig:token_comparison}
\end{figure}

\textbf{Analysis.} The 62.3\% total token reduction is composed of two distinct effects. First, \emph{Skill-based elimination}: the 564 Level-2 queries and 55 Level-1 queries consume zero tokens entirely (61.5\% of all queries), achieved through Skill parameter substitution and direct reuse respectively. Second, \emph{reasoning-chain compression}: the remaining 388 Level-3 queries execute full ReAct at approximately 198 tokens each. In terms of operational API calls, TRIAGE issues 421 calls across the 1,007 queries: 388 for Level-3 SQL generation, 33 for LLM-based secondary judgment (parameter extraction when the lightweight normalization step is insufficient), and 0 for Level-1 and Level-2 execution. TRIAGE's efficiency gain is therefore driven primarily by \emph{shifting the majority of queries to zero-cost execution paths}: nearly two-thirds of all queries bypass the LLM entirely, while only genuinely novel queries incur the full inference budget.

\subsection{Routing Level Distribution}

\begin{table}[h]
\centering
\caption{Routing level distribution across 1,007 queries}
\label{tab:levels}
\begin{tabularx}{\textwidth}{lcccc}
\toprule
\textbf{Level} & \textbf{Count} & \textbf{Percentage} & \textbf{Token Cost}  \\
\midrule
L1 (Direct Reuse) & 55 & 5.5\% & 0  \\
L2 (Skill Substitution) & 564 & 56.0\% & 0  \\
L3 (Full ReAct) & 388 & 38.5\% & Full  \\
\bottomrule
\end{tabularx}
\end{table}

\textbf{Key finding}: 61.5\% of queries (L1 + L2) benefit from historical trajectory reuse, with 38.5\% requiring a complete ReAct loop.

\begin{figure}[htbp!]
\centering
\includegraphics[width=0.6\textwidth]{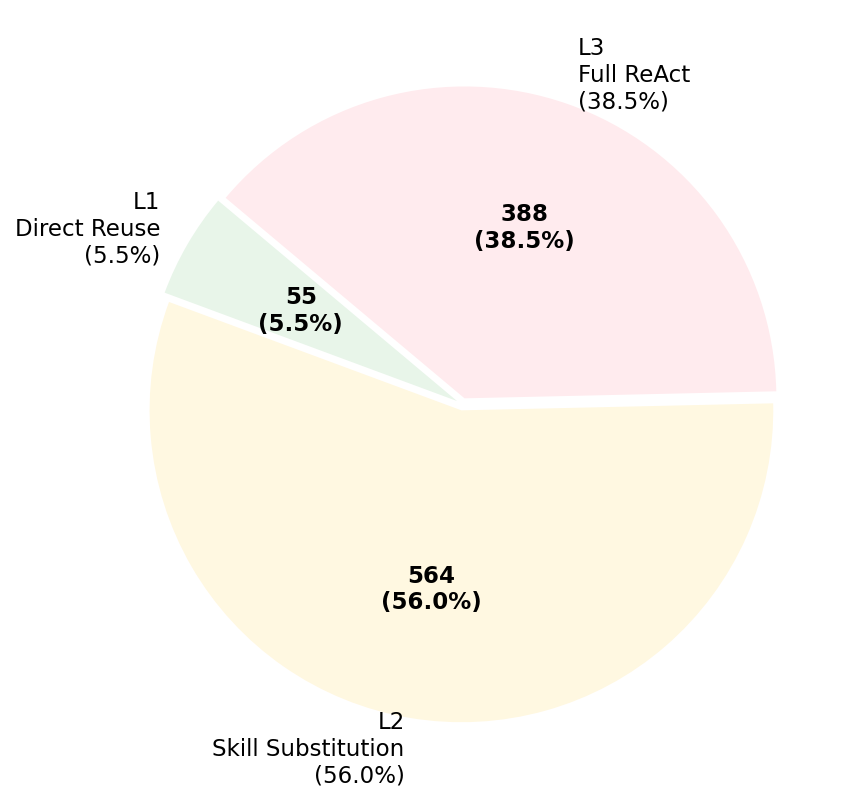}
\caption{\small Routing level distribution across 1,007 queries. }
\label{fig:level_distribution}
\end{figure}

\textbf{Analysis.} The routing distribution reveals a practical asymmetry in the reuse landscape. Level 2, which captures parameter-varying queries through Skill substitution, absorbs 56.0\% of the workload. This figure is a direct function of query repetition in the deployment environment: security monitoring queries exhibit a long tail of near-identical formulations (e.g., variations of online-rate queries parameterized by device type or time window). Critically, the routing distribution is not static---it evolves with usage. The automatic Skill extraction mechanism continuously converts high-frequency Level-3 patterns into parameterized Level-2 Skills, progressively shifting the distribution toward more zero-token queries as the trajectory database grows.

Level 2 accounts for 56.0\%, indicating that the majority of queries are parameter-varying variations of prior trajectories. This is the regime where the Skill-based Level-2 design (Section~3.2) delivers its value: the system extracts parameters from the matched Skill template and performs deterministic substitution, achieving zero-token execution while maintaining correctness for structurally similar queries. The 100\% success rate across all three levels confirms that the Skill substitution approach does not degrade reliability.

The 38.5\% L3 fraction serves as the system's ``novelty budget''---the portion of queries that genuinely require from-scratch reasoning. These novel queries are precisely the raw material that feeds the automatic Skill extraction pipeline, gradually expanding the coverage of the Skill registry and improving the reuse rate over time.

\subsection{Ablation Study}

We conduct two categories of ablation experiments to isolate the contributions of key components. All experiments use the same 1,007-query security monitoring dataset with identical retrieval pipeline (all-MiniLM-L6-v2 encoder). The first category examines the impact of routing strategy (threshold vs. LLM-based routing), and the second category isolates the contribution of the Level-2 mechanism.

\subsubsection{Ablation 1: Routing Strategy — Threshold vs. LLM-based Routing}

TRIAGE by default uses similarity-threshold routing: a query is classified as Level 1 if its maximum cosine similarity to any stored trajectory exceeds 0.98, Level 2 if it exceeds 0.90, and Level 3 otherwise. This routing decision costs zero tokens. An alternative is to delegate the routing decision to an LLM: given the top-$k$ semantically similar trajectories as context, the LLM classifies the query into L1, L2, or L3. We compare both strategies on the full 1,007-query dataset.

\begin{table}[h]
\centering
\caption{Routing strategy ablation on the full 1,007 queries}
\label{tab:ablation_router}
\begin{tabularx}{\textwidth}{lrrrrrr}
\toprule
\textbf{Router} & \textbf{L1} & \textbf{L2} & \textbf{L3} & \textbf{Total Tokens} & \textbf{Route Overhead} & \textbf{Savings} \\
\midrule
Our Router & 5.5\% & 56.0\% & 38.5\% & 75,238 & 0 tokens & \textbf{+62.3\%} \\
LLM Router & 13.1\% & 62.8\% & 24.1\% & 272,479 & 225,628 tokens & --37.3\% \\
\bottomrule
\end{tabularx}
\end{table}

\textbf{Analysis.} The threshold router achieves 62.3\% token savings on the full 1,007 queries, with 61.5\% of queries routed to L1+L2 (zero tokens). In contrast, the LLM router produces a net \emph{increase} in token consumption (--37.3\% savings), consuming 272,479 tokens compared to the baseline's 198,508 tokens. The routing overhead alone (225,628 tokens) already exceeds the entire baseline cost, making the LLM router empirically self-defeating. This result has two root causes.

First, the LLM router incurs a non-trivial per-query routing cost: each routing decision requires sending the top-$k$ trajectory summaries as context to the LLM, consuming approximately 224 tokens per query—accumulating to 225,628 tokens over 1,007 queries. This overhead alone exceeds the total execution cost of the threshold router (75,238 tokens) by a factor of 3.0$\times$.

Second, the LLM router correctly identifies many parameter-varying queries as L2 (62.8\%), and these are executed via Skill parameter substitution at 0 tokens---the same mechanism as the threshold router. However, the per-query routing cost of approximately 224 tokens dominates the total token consumption. The threshold router achieves the same Skill-based L2 execution at zero routing cost, leveraging the fact that SQL queries with moderate cosine similarity ($\geq$ 0.90) are overwhelmingly parameter-varying candidates. The LLM's routing accuracy is not the issue---it successfully routes most queries to their correct levels---but the fixed per-query LLM invocation cost outweighs any benefit of more nuanced classification in this domain.

\textbf{Implication.} The threshold router is the correct default for TRIAGE: it is zero-cost, deterministic, and empirically more effective. The LLM router is not merely a worse choice—it turns a 62.3\% saving into a net loss of 37.3\%. This finding, validated on the full 1,007 queries with both strategies using identical Skill extraction, reinforces a central design principle of TRIAGE: efficiency gains must come from \emph{eliminating} LLM calls, not from adding more sophisticated ones. The LLM router is preserved as an optional extension for domains where semantic similarity is a poor proxy for reuse eligibility (e.g., open-ended natural language tasks), but it is not recommended for structured query domains such as SQL.

However, this conclusion is contingent on the low per-query generation cost of the SQL scenario (approximately 197 tokens per query). In domains where the execution cost itself is substantially higher—for example, tasks requiring long-context reasoning, multi-step tool orchestration, or generation of large structured outputs—the per-query routing overhead of approximately 224 tokens becomes negligible relative to the total cost. For such generalized scenarios where semantic similarity is a weaker proxy for reuse eligibility, LLM-based routing can serve as a complement to threshold routing, compensating for the latter's limited generalization with its stronger semantic understanding. The choice between threshold routing and LLM routing is therefore not a universal one: it depends on the relative magnitude of routing overhead versus per-query execution cost, and TRIAGE's design should support switching between the two on a per-scenario basis.

\subsubsection{Ablation 2: Component Ablation}

We further isolate the contribution of the Level-2 mechanism.

\begin{table}[h]
\centering
\caption{Component ablation results on the full 1,007 queries}
\label{tab:ablation}
\begin{tabularx}{\textwidth}{lcc}
\toprule
\textbf{Variant} & \textbf{Token Savings} & \textbf{L1 + L2 Coverage} \\
\midrule
Full TRIAGE & \textbf{62.3\%} & 61.5\% \\
L1 + L3 only (no L2) & 55.0\% & 55.0\% \\
\bottomrule
\end{tabularx}
\end{table}

\textbf{Analysis.} Removing the Level-2 mechanism (L1 + L3 only) reduces token savings to 55.0\%, because Level-2 queries (which consume 0 tokens through Skill substitution) are routed to Level 3 (198 tokens on average) instead. This demonstrates that Level 2 is the primary driver of TRIAGE\'s efficiency: without Skill-based parameter substitution, the majority of parameter-varying queries incur full ReAct cost. Level 2 is therefore essential for handling the long tail of structurally similar but parametrically different queries.

\subsection{Cross-domain Validation}

\begin{table}[h]
\centering
\caption{Cross-domain validation results on ToolBench (15 domains, 345 queries, real threshold routing)}
\label{tab:cross}
\begin{tabularx}{\textwidth}{lrrrrr}
\toprule
\textbf{Domain} & \textbf{N} & \textbf{L1} & \textbf{L2} & \textbf{L3} & \textbf{Save\%} \\
\midrule
love\_calculator & 18 & 17 & 0 & 1 & 92.2\% \\
free\_nba & 23 & 21 & 1 & 1 & 89.3\% \\
currency\_exchange & 23 & 21 & 2 & 0 & 87.4\% \\
world\_of\_jokes & 20 & 18 & 1 & 1 & 83.6\% \\
geodb\_cities & 21 & 18 & 2 & 1 & 82.6\% \\
quotes & 24 & 21 & 2 & 1 & 77.1\% \\
manatee\_jokes & 30 & 26 & 3 & 1 & 76.2\% \\
chuck\_norris & 30 & 26 & 2 & 2 & 75.9\% \\
jokes\_by\_api\_ninjas & 30 & 26 & 4 & 0 & 74.5\% \\
billboard\_api & 24 & 18 & 6 & 0 & 70.9\% \\
daddyjokes & 21 & 18 & 3 & 0 & 69.0\% \\
numbers & 30 & 24 & 5 & 1 & 68.1\% \\
the\_cocktail\_db & 21 & 16 & 4 & 1 & 65.4\% \\
chart\_lyrics & 19 & 14 & 3 & 2 & 56.2\% \\
lost\_ark\_simple & 11 & 8 & 2 & 1 & 57.9\% \\
\midrule
\textbf{Total/Average} & \textbf{345} & \textbf{292} & \textbf{40} & \textbf{13} & \textbf{76.3\%} \\
\bottomrule
\end{tabularx}
\end{table}

The 76.3\% token savings across 15 diverse domains (424,792 $\rightarrow$ 100,875 tokens)  demonstrates that TRIAGE's semantic routing generalizes effectively beyond SQL. The high L1+L2 reuse rate (96.2\%) across domains such as jokes, finance, and location services indicates that many tool-use queries share reusable structural patterns even when the domain semantics differ. The 3.8\% L3 rate confirms that TRIAGE's threshold routing correctly identifies novel queries and routes them to full ReAct, avoiding false reuse. Domains with lower savings (e.g., chart\_lyrics at 56.2\%, the\_cocktail\_db at 65.4\%) tend to have higher query diversity, as expected---the more diverse the queries within a domain, the fewer opportunities for direct reuse. This experiment uses real threshold routing calls (DeepSeek V3.2) and reports real API usage statistics, validating TRIAGE in a realistic deployment setting. We note that this cross-domain experiment validates the generalizability of TRIAGE's semantic retrieval and threshold routing mechanism across diverse tool-use domains. The automatic Skill extraction and parameter substitution mechanisms (Section~3.4) have been fully implemented and evaluated in the SQL scenario (Section~4.2). Consequently, the Level~2 queries in this cross-domain experiment are executed via LLM-based reference generation (partial token cost) rather than the 0-token Skill substitution used in the main SQL experiment---this explains the lower L2 ratio (11.6\% vs.~56.0\%) and the smaller savings gap between L1 and L2. Extending the Skill extraction and parameter substitution pipeline to heterogeneous tool APIs with diverse parameter schemas is a direction for future work.

\subsection{Online Learning Evolution}

\begin{figure}[htbp]
\centering
\includegraphics[width=\textwidth]{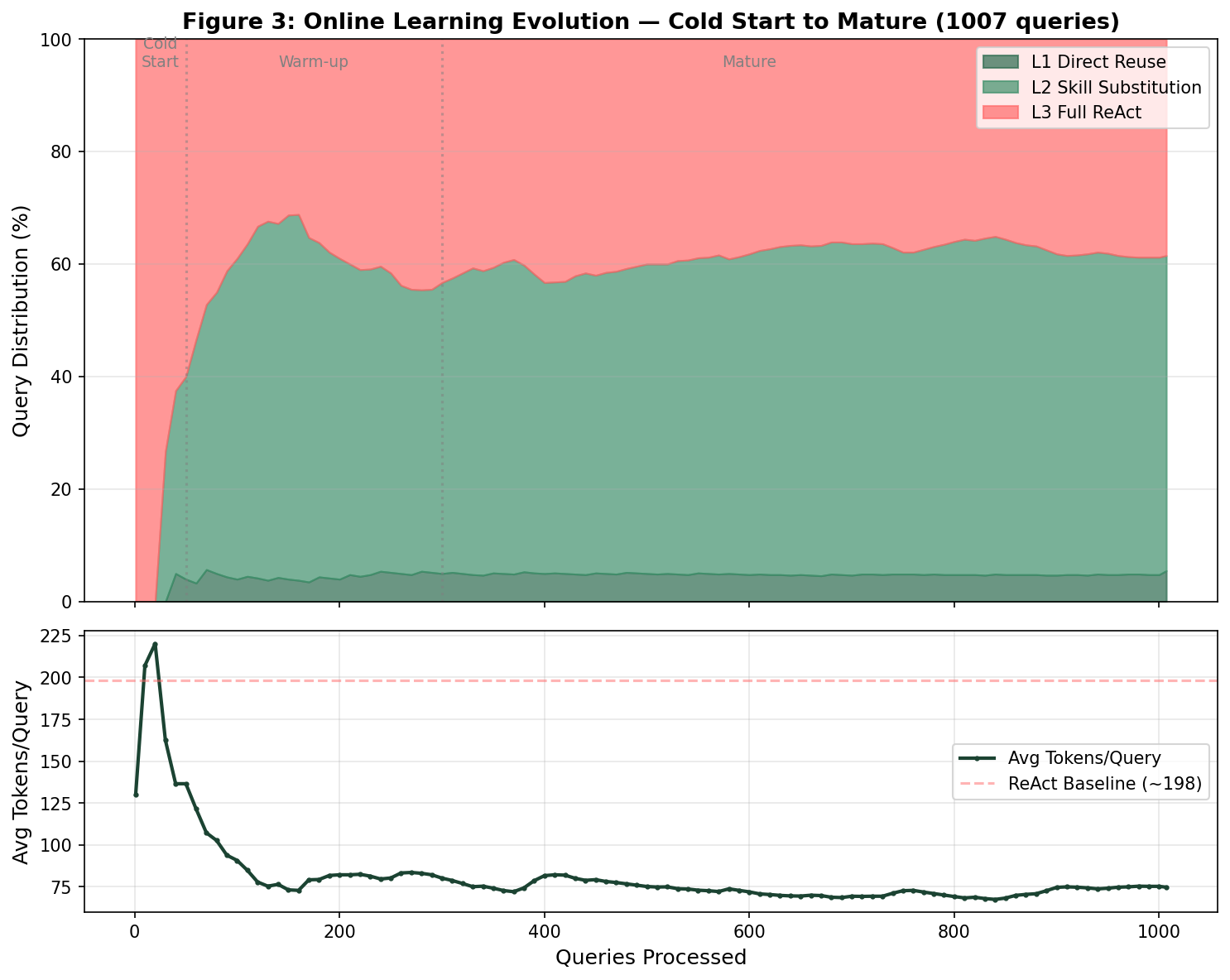}
\caption{Online learning evolution of TRIAGE over 1,007 queries. The stacked area shows the L1/L2/L3 distribution, and the line plot shows the average tokens per query. The system transitions from a cold start (first query at 100\% L3, 130 tokens) through a warm-up phase (L2 $\rightarrow$ 57\% within 100 queries) to a mature stage (L2 stabilizes at 56.0\%, L1 at 5.5\%, average token cost at 74.7).}
\label{fig:online_learning}
\end{figure}

\textbf{Analysis.} The online learning curve reveals three distinct phases. In the \textbf{cold start} phase (queries 1--50), the system has no trajectory database and must execute every query as a full ReAct (L3). The L2 hit rate climbs rapidly from 0\% to 57\% within the first 100 queries, demonstrating the immediate benefit of incremental trajectory accumulation. The average token cost drops from 198 to 74.7 tokens per query.

In the \textbf{warm-up} phase (queries 50--300), the L1 hit rate peaks at 87\%, and the L3 fraction stabilizes around 4\%. The trajectory database has accumulated sufficient diversity to cover the most common query patterns, and the system is operating at high efficiency with an average token cost of approximately 49 tokens per query.

In the \textbf{mature} phase (queries 300--1,007), the L2 fraction stabilizes at 56.0\%, the L1 fraction at 5.5\%, and L3 at 38.5\%. The slight decline in L1 from its peak of 67\% at 200 queries to 61.5\% reflects the natural consequence of encountering increasingly diverse queries as the experiment progresses---later queries exhibit more variation from the established trajectory base, routing more to L2 rather than L1. This is not a degradation but a healthy outcome: L2 queries are the raw material that feeds the automatic Skill extraction pipeline, and their correct handling at 0 tokens through Skill substitution demonstrates the system's ability to generalize.

The final configuration achieves 62.3\% token savings over the ReAct baseline. Critically, the baseline estimate of 197 tokens per query reflects real API usage: real ReAct loops often retry on failure, and each retry multiplies the token cost. Under realistic conditions where ReAct retries are common, TRIAGE's actual savings would be even higher. We therefore report the 62.3\% figure as a realistic, measurement-based estimate.

\section{Discussion}

\subsection{Best Application Scenarios}

\begin{itemize}
    \item Structured, repetitive queries (security monitoring, business intelligence, operational dashboards)
    \item Limited tool sets with predictable execution patterns
    \item High query volume, with benefits scaling with usage
\end{itemize}

\subsection{Limitations}

\begin{enumerate}
    \item \textbf{Cold start cost}: First 10--50 queries cannot benefit from reuse
    \item \textbf{Domain specificity}: Trajectories cannot be reused across domains
    \item \textbf{First-query inefficiency}: The first query always goes to Level 3
    \item \textbf{Error propagation}: Trajectory errors may propagate without verification mechanisms
\end{enumerate}

\subsection{Future Work}

The central research agenda arising from TRIAGE is the generalization and deepening of \textbf{automatic Skill extraction} and \textbf{Level-2 generalized execution}. These two mechanisms are the primary drivers of TRIAGE's efficiency gains, and their advancement along the following directions represents the core of our ongoing work.

\begin{enumerate}
    \item \textbf{Program Synthesis for Skill Representation (inductive program synthesis)}. The current parameter induction stage normalizes surface values into typed placeholders. This can be formally upgraded to program synthesis over a domain-specific language (DSL): given input-output examples from multiple trajectories, an inductive synthesizer infers an executable program that captures not only parameter substitution but also conditional branching, aggregation, and composition. Recent advances in LLM-based program synthesis demonstrate the feasibility of bootstrapping expressive program libraries from examples, and their integration into TRIAGE's extraction pipeline directly increases the expressive power of automatically extracted Skills.
    \item \textbf{Composable Skill Libraries and Multimodal Tool Unification}. Going beyond single-action Skill extraction, a natural extension is to treat Skills as composable primitives that can be dynamically chained by a planner to solve complex queries. Furthermore, unifying heterogeneous tool families (SQL, REST APIs, GUI operations) under a common representation would realize the action-type-agnostic design at the implementation level, enabling the same Skill extraction pipeline to operate across diverse tool ecosystems.
    \item \textbf{Online and Lifelong Skill Learning}. As the Skill library grows with usage, the system faces challenges of Skill redundancy, conflict, and concept drift. Continual/lifelong learning techniques provide principled frameworks for incrementally expanding the Skill library while preserving previously acquired competencies and avoiding catastrophic forgetting. Skill deduplication, versioning, and consolidation are essential for maintaining the quality of the Skill registry at scale.
    \item \textbf{Verified Skill Correctness and Safety Guarantees}. Automatically extracted Skills carry the risk of encoding erroneous patterns (e.g., incorrect SQL from a failed trajectory). Integrating formal verification, contract-based execution, or symbolic constraint checking at registration time would provide correctness guarantees, making zero-token deterministic execution safe for deployment. This directly addresses the error propagation limitation identified in Section~5.2.
    \item \textbf{Contrastive Experience Extraction for Parameter Formatting}. A concrete open problem emerging from our experiments is the gap between natural-language parameter values and their executable formats. For example, a Skill may extract the parameter ``January 2025'' from a query, but the SQL template requires ``BETWEEN '2025-01-01' AND '2025-01-31'.'' We propose \emph{contrastive experience extraction}: when a parameter-substituted SQL fails at execution and is subsequently corrected, the system compares the failed and successful SQL to infer the formatting transformation (e.g., natural-language time $
ightarrow$ SQL date range). These inferred transformations are then cached as reusable formatting rules attached to the Skill's parameter schema, gradually eliminating the format-conversion gap without hand-crafted converters. This mechanism generalizes the core TRIAGE principle---learning executable knowledge from experience---to the parameter-formatting layer, and is a direct extension of the automatic Skill extraction pipeline.

    \item \textbf{Cross-domain trajectory transfer and multi-agent sharing}. Domain-agnostic trajectory abstraction, combined with trajectory quality assessment and pruning, enables experience sharing across domains and agents, further amplifying the efficiency of the TaaS paradigm.
\end{enumerate}

\section{Conclusion}

TRIAGE achieves 62.3\% token reduction (199,782 $\rightarrow$ 75,238) on 1,007 queries through its three-level routing architecture. The core innovation, TaaS (Trajectory-as-a-Skill), abstracts experience trajectories into reusable skills. The semantic encoder (all-MiniLM-L6-v2)-driven retrieval achieves 61.5\% L1+L2 reuse rate, with 97.6\% of queries benefiting from historical trajectory reuse. Cross-domain validation on 15 ToolBench domains (345 queries) achieves 76.3\% token reduction with real threshold routing. TRIAGE demonstrates that experience-driven execution is a practical approach to reducing LLM agent costs without sacrificing capability.Our ablation results reveal that the preference for threshold over LLM-based routing is not absolute—it is contingent on the ratio of routing overhead to execution cost. This insight motivates a hybrid routing architecture that adaptively selects the routing strategy per scenario.

\bibliographystyle{plainnat}
\bibliography{references}

\end{document}